\documentclass[runningheads]{llncs}
\usepackage[T1]{fontenc}
\usepackage{graphicx,verbatim}
\usepackage{svg}
\usepackage{amsmath,amssymb}
\usepackage{soul}
\usepackage{ulem}
\usepackage{xspace}
\usepackage[dvipsnames]{xcolor}
\usepackage{pgfplotstable}

\newcommand{\ours}{MeiBRD\xspace}
\begin{document}

\title{MeiBRD: Meta-Learning Intraoperative Biomechanical Residual Deformation}
\author{Casey Meisenzahl\inst{1} \and
Jon Heiselman\inst{2} \and
Michael Holtz\inst{1} \and
Yubo Ye\inst{1} \and
Michael Miga\inst{2} \and
Linwei Wang\inst{1}}
\authorrunning{C. Meisenzahl et al.}
\institute{Rochester Institute of Technology, NY, USA \and
Vanderbilt University, Nashville TN, USA}

\maketitle
\begin{abstract}
Accurate intraoperative liver registration is challenging due to substantial soft-tissue deformation yet sparse intraoperative measurements. 
Biomechanical models regularize this ill-posedness with prior knowledge but exhibit persistent prediction bias due to simplifying assumptions, while data-driven learning solutions struggle with data efficiency, generalization, and physical plausibility.
We propose a hybrid registration framework that adapts a biomechanical prior using sparse intraoperative correspondences. 
Rather than learning a full deformation field, we learn a residual deformation function that corrects linear biomechanical predictions, 
modeled as a graph neural diffusion function with geometry-aware attention over the 3D liver mesh.  
To enable long-range information transfer of 
sparse observations, 
we take a novel perspective 
of sparse intraoperative measurements 
as \textit{context} samples where 
input-output pairs of 
the residual deformation function are fully observed, 
casting the problem into 
learning-to-learn this residual function from intraoperative context samples with feedforward meta-learners.
Experiments on a deformable liver phantom dataset demonstrate improved registration accuracy and generalization compared to rigid, biomechanical, and data-driven baselines, particularly for out-of-distribution geometries and deformations.


\end{abstract}
\section{Introduction}
During laparoscopic liver resections
 \cite{abuhilal_southampton_2018}, 
the liver undergoes substantial intra-procedural deformation due to factors such as patient positioning, pneumoperitoneum, respiration, and tool–tissue interaction, rendering preoperative imaging alone insufficient to accurately represent the intraoperative anatomy.
Image-to-physical registration aims to resolve this problem by estimating a spatial transformation that aligns preoperative imaging data with the patient’s intraoperative physical anatomy.
The associated registration problem unfortunately is significantly ill-posed because measurements of the intraoperative anatomy are often only available at very sparse locations \cite{miga_computational_2016}, 
acquired for instance with tracked stylus 
covering 15-25\% of the liver surface or several tracked intraoperative ultrasound (iUS) planes providing sparsae surface and subsurface data.

Biomechanical based registration has been an important approach to regularize this ill-posed problem by constraining deformations to be physically plausible \cite{brunet_physicsbased_2019}. 
By enforcing the governing equations of soft-tissue mechanics, deformations are iteratively estimated to minimize model-data mismatch under sparse intraoperative observations \cite{heiselman_intraoperative_2020}.
However, to remain computationally tractable and numerically well-conditioned for intraoperative use, these approaches typically rely on simplifying assumptions, such as linear elasticity and approximate boundary conditions, 
which limits their ability to capture lage and nonlinear deformations. 
 
In recent years, deep learning (DL) based deformable registration has been explored to model complex, nonlinear tissue behavior directly from data \cite{pfeiffer_nonrigid_2020}. While these approaches offer fast inference and expressive modeling capacity, their performance in the specific context of image-to-physical registration is limited by the scarcity of ground-truth deformation data and the sparsity of intraoperative measurements, leading to poor generalization outside the training distribution and physically implausible predictions \cite{heiselman_comparison_2023}. 
To address this, 
a hybrid approach was recently proposed to learn data-driven corrections to 
predictions made by linear biomechanical models, 
demonstrating improved registration accuracy compared to fully biomechnaics- or DL-based approaches
 \cite{wang_libr_2024}. However, whether learning full deformations or residuals, existing neural approaches typically treat sparse intraoperative measurements as node-wise features that must be propagated across the entire liver mesh.  
 In this formulation, 
 the sparsity of intraoperative measurements  
 becomes a critical bottleneck for  
 an effective long-range information transfer throughout the entire liver anatomy to be deformed.

\begin{figure}[t]
    \centering
    \includegraphics[width=0.9\linewidth]{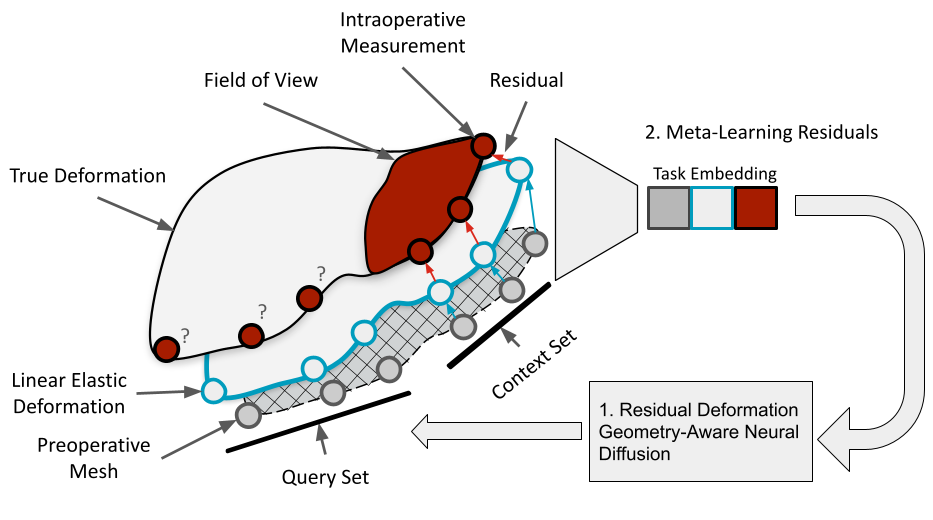}
    \caption{Our method has two main components: (1) modeling residuals of linear biomechanical deformation as a  geometry-aware diffusion function,
  and (2) a meta-learning formulation of learning-to-learn this residual function from sparse intraoperative correspondences as context samples with rapid feedforward meta-learners. 
  }
    \label{fig:overview}
\end{figure}

To address these challenges, we 
present \ours (Meta-Learning Intraoperative Biomechanical Residual Deformation) 
to recast the problem of image-to-physical registration as \textit{meta-learning a function describing the \textit{local} residual of linear biomechanical deformation,} considering intraoperative measurements as samples where
this function is fully observed. 
As illustrated in Fig.~\ref{fig:overview}, 
we first model this deformation residual function $f_\text{res}$ as a 
continuous neural diffusion process defined over 
the geometrical graph of the liver, 
with novel strategies to injest structural information of the liver geometry into the diffusion operator. 
Interpreting locations with intraoperative measurements as observed samples of this residual function where the inputs (pre-operative geometry and deformation predicted by linear biomechanical model) and output (residual to true deformations) are available, 
we then cast the problem of residual estimation 
in a meta-learning framework 
where locations with intraoperative measurements are used as \textit{context} samples, to adapt the biomechanical residual function $f_\text{res}$ to predict the residual for the rest of the liver. 
Given time constraints in surgical environments, 
this is achieved with a feedforward meta-learner 
where the extraction of information from intraoperative measurements and the adaption of the residual function is all done with rapid feedforward inference. 
We evaluated \ours on a phantom  liver dataset \cite{heiselman_intraoperative_2020}, comparing against rigid ICP, biomechanical registration \cite{heiselman_intraoperative_2020,yang_boundary_2024}, and data-driven volume-to-surface registration \cite{pfeiffer_nonrigid_2020}, demonstrating consistent improvements in registration accuracy and generalization.

\section{Methodology}

Consider a preoperative liver geometry represented as a mesh graph
$\mathcal{M} = (V, E)$ with $n$ vertices
and edges 
\( E \subseteq V \times V \) defined by the tretrahedron mesh. 
Each vertex is associated with a 3D coordinate 
\( \mathbf{x}_i \in \mathbb{R}^3 \). 
Stacking all vertex coordinates yields the mesh configuration 
$
X = 
\begin{bmatrix}
\mathbf{x}_1^\top & \dots & \mathbf{x}_n^\top
\end{bmatrix}^\top
\in \mathbb{R}^{3n}.$
Let $Y^* \in \mathbb{R}^{3n}$ denote the true intraoperative mesh vertex configuration. 
Assume we have 1) a set of intraoperative measurements represented by $Y^{\mathrm{msr}} \in \mathbb{R}^{3l}$ with $l<<n$, 
and 2) a deformation predicted by fitting a linear-elastic biomechanics model to the sparse measurements, represented by $X_{\mathrm{lin}} \in \mathbb{R}^{3n}$. 
Our goal is to learn a residual correction $\mathbf{r}_i \in\mathbb{R}^{3}$ between the biomechanically-predicted deformation and the truth given by $\mathbf{y}_i^* = \mathbf{x}_{\mathrm{lin},i} + \mathbf{r}_i$ for all vertices $i$ in the mesh graph $\mathcal{M}$. 
To this end, \ours consists of two key elements as outlined in 
Fig.~\ref{fig:overview}. First, we model the residual of linear biomechanical deformation $\mathbf{r}$ at vertex $i$ as a continuous graph neural diffusion (GRAND) process
$\mathbf{r}_i = f_\text{res}(\mathbf{z}^{\mathcal{N}(i)},\mathbf{e}^{\mathcal{N}(i)})$, 
where $\mathbf{z}$ and $\mathbf{e}$ denote novel geometrical node and edge features 
with the neighborhood $\mathcal{N}(i)$ defined by mesh $\mathcal{M}$.  
Second, 
we consider sparse intraoperative measurements as observed samples of  $\mathbf{r}$ and approach intraoperative liver registration from meta-learning how to adapt $f_\text{res}$ using observed context examples.

\subsection{Biomechanical Residual as Geometry-Aware Neural Diffusion}
\label{subsec:grand}

Given the mesh graph
$\mathcal{M} = (V, E)$,
we model the residual of linear biomechanical deformation at vertex $i$, $\mathbf{r}_i = f_\text{res}(\mathbf{z}^{\mathcal{N}(i)},\mathbf{e}^{\mathcal{N}(i)})$, with GRAND \cite{chamberlain_grand_2021} as:
\begin{eqnarray}
\nonumber
f_\text{res} &= 
l_\theta \big(
 \mathbf{z}(0) + \int_{s=0}^1 \frac{d\mathbf{z}(t)}{dt} ds \big), \quad  
\frac{d\mathbf{z}(t)}{dt} =
\bigl(\mathbf{A}(\mathbf{z}(t))-\mathbf{I}\bigr)\,\mathbf{z}(t), \\
A(i,j) &=\mathrm{softmax}_{j\in\mathcal{N}(i)}
\Bigl(\tfrac{(\mathbf{W}_Q\,\mathbf{z}_i)^\top (\mathbf{W}_K\,\mathbf{z}_j)}{\sqrt{d}}\Bigr),
\qquad (i,j)\in E,
\label{eq:grand_u_ode}
\end{eqnarray}
where $\mathbf{z}(0)$ is the input initial node embedding on V, 
$\mathbf{A}(\cdot)$ is an adjacency-structured attention operator, 
$\mathbf{W}_Q$ and $\mathbf{W}_K$ are query and key projection matrics, 
and $d$ is the dimention of the attention head.
$l_\theta$ is an MLP paraemterized by $\theta$ that maps 
the output of GRAND's ODE solver to  output residual $\mathbf{r}$.

\subsubsection{Incorporating geometrical features:} 
We enrich $f_\text{res}$ with geometrical features $\mathbf{g}$ that are known to be important for soft-tissue deformation. 

For surface geometry, 
we consider the mean surface curvature 
as a compact descriptor of local bending 
where high-curvature regions tend to produce large non-linear shape changes with small loads/constraints \cite{Meyer2003DDG}. 
For each surface vertex $i$, 
we define
$\mathbf{g}_i^\text{surf} = \frac{1}{2A_i}\sum_{j\in \mathcal{N}^s(i)}
(\cot\alpha_{ij}+\cot\beta_{ij})(\mathbf{x}_i-\mathbf{x}_j)$,
where  $\mathcal{N}^s(i)$ defines the 1-ring surface neighbors for vertex $i$, 
$\alpha_{ij}$ and $\beta_{ij}$ are the angles opposite edge $(i,j)$ in the two incident surface triangles, 
and $A_i$ the barycentric area at $i$.

For volume geometry, 
we consider volume changes of tetrehedron elements which capture local compression/expansion as a primary driver of nonlinear bulk deformation \cite{BonetWoodNonlinearContinuumMechanics}. 
For each internal vertex $i$, 
we define 
$\mathbf{g}_i^\text{vol} = \frac{1}{K} \sum_{
    i \in V_k
} \det(\mathbf{F}_k)$ 
where $V_k$ represents
vertices of tetrahedron element $k$, 
$\mathbf{F}_k$ its gradient of linear biomechanical deformation, 
and $K$ the number of elements vertex $i$ belongs to. 
For all vertices, 
we further consider distortional (volumetric) strain features 
as a rotation-invariant measure of an element's finite extensional strain \cite{Belytschko2013NonlinearFEM}: 
by reflecting the magnitude of stretch, this additional feature can distinguish deformations with similar volume changes but different stretching patterns \cite{BonetWoodNonlinearContinuumMechanics}.
For each vertex $i$, 
we define 
$\mathbf{g}_i^\text{strain} = \frac{1}{K} \sum_{
    i \in V_k
} \operatorname{tr}(\mathbf{E}_k)$ 
where $\mathbf{E}_k$ is the
Green--Lagrange strain tensor 
of linear biomechanical deformations
 at tetrahedron element $k$.

With this, we consider two types of node features. 
Surface node features represent coordinate and surface curvature changes of the liver
due to linear biomechanical deformations, along with its volumetric strain as:
\begin{equation}
\label{eq:geo_surf}
    \mathbf{x}^\text{surf}_g = [\mathbf{x}, \mathbf{x}_\text{lin},\nabla \mathbf{x}, \mathbf{g}^\text{surf}, 
 \mathbf{g}^\text{surf}_\text{lin}, 
 \nabla \mathbf{g}^\text{surf}, \mathbf{g}^\text{strain}
]
\end{equation} 
Internal node features represente coordinate and volume changes due to linear biomechanical deformations, along with its volumetric strain as:
\begin{equation}
\label{eq:geo_vol}
    \mathbf{x}^\text{vol}_g = [\mathbf{x}, 
\mathbf{x}_\text{lin}, \nabla \mathbf{x}, \mathbf{g}^\text{vol}, \mathbf{g}^\text{strain}
]
\end{equation}

We further define edge features $\mathbf{e}_{ij}$ to capture the relative position between vertices $i$ and $j$ both pre and post the linear biomechanical deformation:
\begin{equation}
\label{eq:geo_edge}
    \mathbf{e}_{ij} = [\mathbf{x}_i - \mathbf{x}_j, 
\mathbf{x}_{\text{lin},i} - \mathbf{x}_{\text{lin},j}]
\end{equation}

\subsubsection{Geometry-aware diffusion:}
To incorporate the geometrical node features defined 
in Equations \eqref{eq:geo_surf}-\eqref{eq:geo_vol} into the 
diffusion operator \eqref{eq:grand_u_ode}, 
we simply define two node-embedding networks, parameterized by 
$\phi_s$ and $\phi_v$, respectively  
for extract input node-embedding $\mathbf{z}(0)$ at surface and internal vertices:
\begin{equation}
\label{eqn:embed}
    \mathbf{z}(0) = 
    \begin{cases}
     g_{\phi_s}(\mathbf{x}_g^{\text{surf}}), & \text{for surface vertices}  \\
      g_{\phi_v}(\mathbf{x}_g^{\text{vol}}), & \text{for internal vertices}
      \end{cases}
\end{equation}

To capture potential directional anisotropic deformation, we further introduce an edge-embedding 
$b_\psi(\mathbf{e}_{ij})$ that 
maps edge features $\mathbf{e}_{ij}$, the relative positions between two vertices, into a scalar that bias the attention score $A_{ij}$
as:
\begin{equation}
\label{eq:grand:bias}
    A(i,j)=\mathrm{softmax}_{j\in\mathcal{N}(i)}
\Bigl(\tfrac{(\mathbf{W}_Q\,\mathbf{z}_i)^\top (\mathbf{W}_K\,\mathbf{z}_j)}{\sqrt{d}} + b_\psi(\mathbf{e}_{ij})\Bigr),
\qquad (i,j)\in E,
\end{equation}
where $b_\psi$ is obtained via an MLP with parameter $\psi$ 
and expected to learn to suppress or enhance attention depending on the direction of an edge.  

\subsection{Meta-Leaerning Biomechanical Residual Functions}
\label{subsec:meta}

Function $f_\text{res}$ defined in Section \ref{subsec:grand} models the residual of linear biomechnical deformations as a function of local geometrical features on pre-operative and linearly-deformed liver meshes. 
We now take a novel perspective 
and view the locations with intraoperative measurements 
as locations where the input-output pairs of $f_\text{res}$ are observed. 
To this end, 
we first estalish this input-output correspondence 
and then formulate the problem of optimizing $f_\text{res}$ as a problem of 
learning-to-learn $f_\text{res}$ from these sparse context examples.

\subsubsection{Correspondence as context examples:}
Since sparse measurements $Y^{\mathrm{msr}}$ represent a subset of true deformations, we use these samples to construct residual pairs with the deformed mesh $X_{\mathrm{lin}}$. These pairs are realizations of the residual function we aim to approximate. For each measurement in $Y^{\mathrm{msr}}$, we identify the nearest neighboring vertex in $X_{\mathrm{lin}}$, to form $
\mathcal{C} = \{(\mathbf{x}^{\mathrm{msr}}_{\mathrm{lin},i}, 
\mathbf{r}^{\mathrm{msr}}_i :=
\mathbf{y}^{\mathrm{msr}}_i - \mathbf{x}^{\mathrm{msr}}_{\mathrm{lin},i})\}_{i=1}^l$. This correspondence is constrained by the acquisition modality: sparse measurements collected with the tracked stylus are associated with the exterior liver surface, while points collected from iUS are associated with subsurface veins.

\subsubsection{Feedforward meta-learning of $f_\text{res}$:}
While gradient-based meta-learning, 
such as MAML \cite{Finn2017MAML}, 
is straightforward and widely used, 
it requires backpropagation of gradients at the time of adaptation as context samples become available. 
To meet the time constraint in intraoperative environment, 
we formulate a feedforward meta-learner to 
obtain 
$f_\text{res}$ from $Y^\text{msr}$ as they become available. This meta-learner consists of two key components: 
1) a meta-encoder that extracts context embeddings from available $
\mathcal{C} = \{(\mathbf{x}^{\mathrm{msr}}_{\mathrm{lin},i}, \mathbf{r}^{\mathrm{msr}}_i)\}_{i=1}^l$, 
and 
2) a hypernetwork that generates the key parameters of $f_\text{res}$ from this context embedding. 

Given
$\mathcal{C} = \{(\mathbf{x}^{\mathrm{msr}}_{\mathrm{lin},i}, \mathbf{r}^{\mathrm{msr}}_i)\}_{i=1}^l$, 
we first extract the corresponding node embeddings $\{\mathbf{z}^{\mathrm{msr}}_{i}\}_{i=1}^{l}$ with the same embedding networks as defined in Equation \eqref{eqn:embed}. 
To fuse the information between $\{\mathbf{z}^{\mathrm{msr}}_{i}\}_{i=1}^{l}$  and their corresponding measured residuals  $\{\mathbf{r}^{\mathrm{msr}}_{i}\}_{i=1}^{l}$ across the context set, 
we define:
\begin{equation}
    \omega = 
\sum\nolimits_{i=1}^K
g_\varphi(\mathbf{z}^{\mathrm{msr}}_{i} +
g_\xi(\mathbf{r}^{\mathrm{msr}}_{i})),
\end{equation}
where $g_\xi$ is an up-projection and 
$g_\varphi$ a down-projection parameterized by $\xi$ and $\varphi$ respectively.
We then use $\omega$ to generate the query/key projection matrices in Equation \eqref{eq:grand:bias} with a hyper-network $h_\zeta$ parameterized by $\zeta$:
\begin{equation}
(\mathbf{W}_Q, \mathbf{W}_K) = h_\zeta(\omega),    
\end{equation}
such that $f_\text{res}$ becomes conditioned on available intraoperative measurements as: 
$\mathbf{r}_i = f_\text{res}(\mathbf{z}^{\mathcal{N}(i)},\mathbf{e}^{\mathcal{N}(i)}; \mathcal{C})$. 
The learning-to-learn $f_\text{res}$ objective can then be formulated as 
asking the context-conditioned $f_\text{res}(\mathbf{z}^{\mathcal{N}(i)},\mathbf{e}^{\mathcal{N}(i)}; \mathcal{C})$ 
to estimate the residual of linear deformation 
at the rest of the mesh locations, $\mathbf{x}_{\text{lin},i} \notin \mathcal{C}$, where intraoperative measurements are not available: 
\begin{equation}
\label{eqn:meta:q}
\mathcal{L}^q=
\sum\nolimits_{\mathbf{x}_{\text{lin},i} \notin \mathcal{C} }
   || \mathbf{y}_i^* - (\mathbf{x}_{\mathrm{lin,i}} + f_\text{res}(\mathbf{z}^{\mathcal{N}(i)},\mathbf{e}^{\mathcal{N}(i)}; \mathcal{C} ) ||_2^2
\end{equation}

Given $N$ samples of 
$\{ X_\text{pre}^j, X_\text{lin}^j, Y^{*,j}, Y^{\text{mrs},j}\}_{j=1}^N$ 
with $\mathcal{L}^{q,j}$ defined for each $j$-th sample, 
we optimize \ours as:
\begin{equation}
\label{eqn:meta:all}
    \hat{\Theta} = \arg\min_\Theta \sum\nolimits_{j=1}^N \mathcal{L}^{q,j}, \qquad \text{where} \quad \Theta = \{\theta, \phi_s, \phi_v, \psi, \varphi, \xi, \zeta \}
\end{equation}
which can be viewed as a meta-objective where 
$\mathcal{L}^{q,j}$ in Equation \eqref{eqn:meta:q} defines the query loss on one \textit{task} which is then aggregated across $N$ tasks in Equation \eqref{eqn:meta:all}.

\section{Experiments and Results}

\textbf{Data:} Experiments were conducted using the phantom liver dataset described in \cite{heiselman_intraoperative_2020}. The dataset comprises three unique geometries, each subjected to three deformation conditions: left mobilization (L), no mobilization (N), and right mobilization (R), corresponding to different ligament dissections. In total, this yields nine deformed phantoms. 
For each deformation, 700 intraoperative scenarios were generated with varying levels of sparse measurements, including approximately 15–25\% of the liver surface covered by a tracked stylus and combinations of one to three 16-plane iUS acquisitions.

\textbf{Baselines:} 
We considered representative existing works in the space of image-to-physical registration designed to handle sparse intraoperative measurements, 
including 
rigid ICP,
biomechnical-model based registraion using 
linear-elastic models (LIBR) \cite{heiselman_intraoperative_2020} 
or nonlinear models (BCF-FEM) \cite{yang_boundary_2024}, and data-driven volume-to-surface registration (V2S) \cite{pfeiffer_nonrigid_2020}. 
While the residual approach presented in \cite{wang_libr_2024} was the closest to \ours, 
we were unable to include it as a baseline due to insufficient details for faithful reproduction.

\begin{table}[t]
\centering
\caption{Target registration error (TRE) of all models. Bold highlights best performance, with the second-best accuracy within statistical significance underlined.}
\begin{tabular}{lccc}
\hline
Method & Random Scenario & Geometry-Deformation & Geometry \\
\hline
wICP      & 13.84 (8.92) & 13.86 (8.39) & 13.86 (8.39) \\
LIBR \cite{heiselman_intraoperative_2020}    & 6.71 (4.92)  & 6.66 (4.61)  & \uline{6.66 (4.61)}  \\
BCF-FEM \cite{yang_boundary_2024}  & 10.43 (7.68) & 10.43 (7.68) & 10.43 (7.68) \\
V2S \cite{pfeiffer_nonrigid_2020}     & \uline{3.63 (3.42)}         & 7.06 (3.82)         & 14.43 (4.34)        \\
MeiBRD   & \textbf{3.56 (2.37)}  & \textbf{5.43 (3.19)}  & \textbf{6.17 (3.66)}  \\
\hline
\end{tabular}
\label{tab:baselines}
\end{table}

\begin{figure}[t]
\centering
\includegraphics[width=1\linewidth]{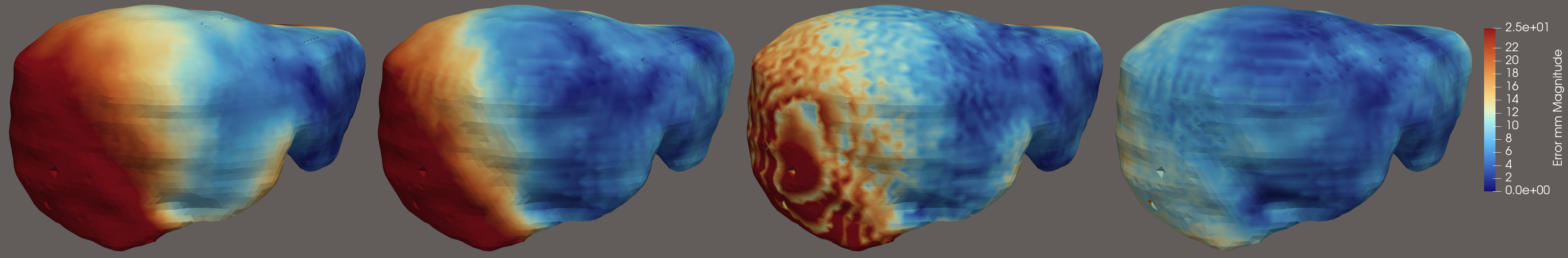}
\caption{Per-vertex error of liver mesh deformation prediction relative to ground truth (blue = low error; red = high error). Left to right: wICP, LIBR, V2S, and MeiBRD. .}
\label{fig:baselines_eg}
\end{figure}
\textbf{Metrics:} 
We assessed three different levels of generalization difficulties. 
In the random split, available intraoperative data scenarios were randomly divided into training, validation, and testing sets. In the geometry–deformation split, data were separated by mobilization condition (L, N, R) across the three phantom geometries (1–3), such that the test set contained unseen deformations of geometries observed during training. In the geometry-based split, partitions were defined by phantom geometry independent of deformation type, with one geometry used for training, one for validation, and one for testing. 
All scenarios were trained and tested in 3-fold cross-validation.
Test errors were quantified using the mean target registration error (TRE) as the average Euclidean distance between corresponding vertices of the estimated and true deformed mesh.

\textbf{Results:} 
Table \ref{tab:baselines} summarizes the quantative results of all models. As shown, 
rigid registration (wICP) and biomechanical-model based approaches 
(LIBR and BCF-FEM) are training-free 
and thus not impacted by out-of-distribution (OOD) scenarios: 
notably, linear biomechanics based LIBR 
demonstrated the best performance among these three approaches,  
indicating the usefulness of simple linear deformation models. 
The data-driven baseline of V2S demonstrated excellent performance in the random split, 
which unfortunately deteriorated rapidly in OOD scenarios, 
providing evidence for the challenge of generalization. 
\ours, in comparison, 
obtained the lowest TRE across all test scanerios and demonstrated robustness to unseen deformations or geometries.
This improvement can be seen in the visual examples in
Figure \ref{fig:baselines_eg}.

\begin{figure}[t]
\centering
\includegraphics[width=0.8\linewidth]{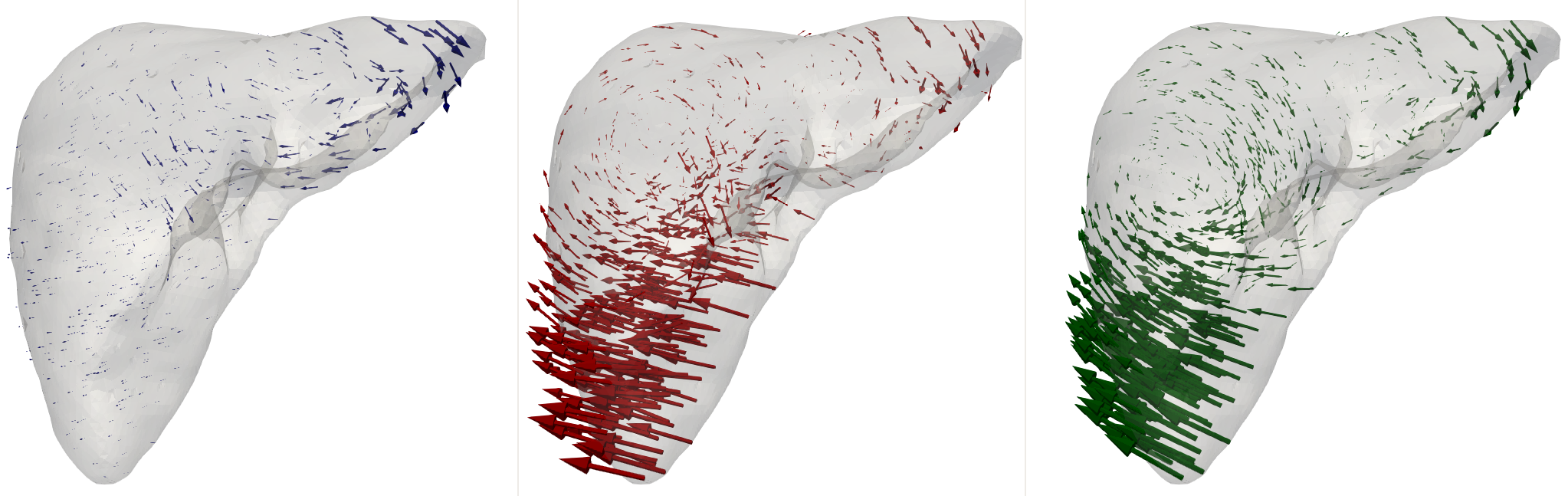}
\caption{Left: deformation field predicted by the biomechanical model (LIBR). Center: ground-truth deformation field. Right: corrected prediction obtained by adding the predicted residual to the biomechanical model deformation field.}
\label{fig:pred_eg}
\end{figure}

\begin{figure}[t]
\centering
\includegraphics[width=0.9\linewidth]{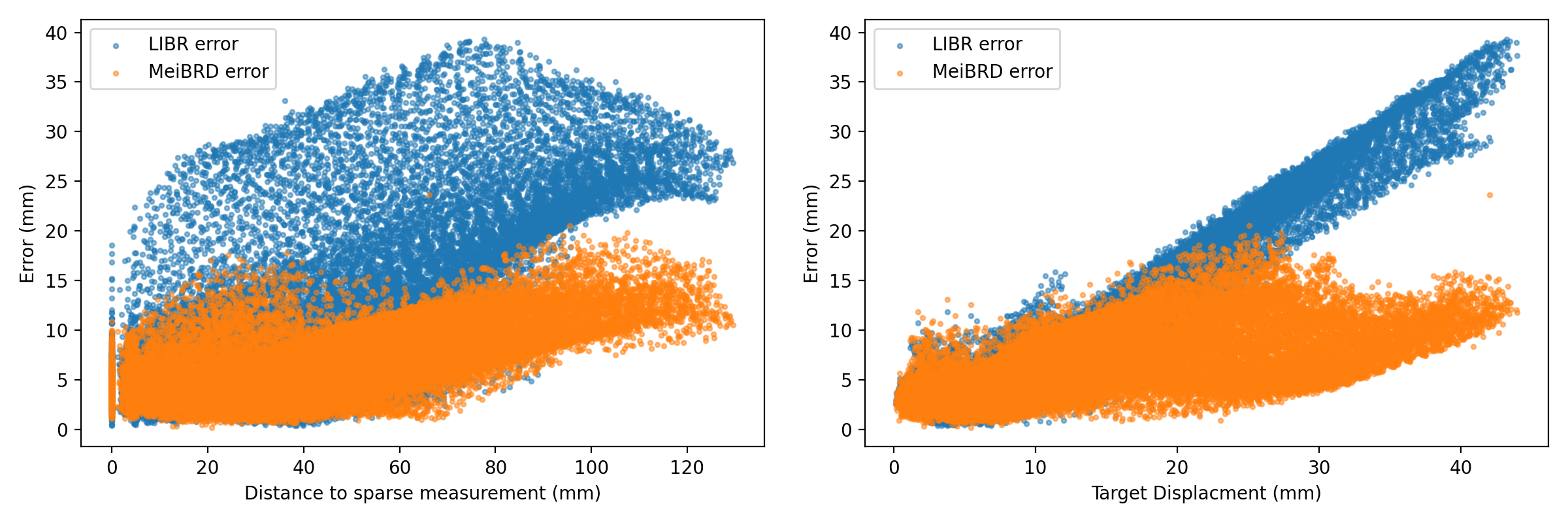}
\caption{(Left) Prediction error (y-axis) as a function of distance to the nearest sparse measurement (x-axis). MeiBRD (orange) maintains low error even at large distances from contextual inputs, indicating effective long-range information transfer. (Right) Prediction error (y-axis) versus true deformation magnitude (x-axis), highlighting performance across increasing large deformations and the ability to resolve residuals in linear biomechanical deformations (LIBR, blue).}
\label{fig:error_wrt}
\end{figure}

Figure \ref{fig:pred_eg} provides another visual example demonstrating how \ours resolves the error in the linear biomechanical deformations. As shown, the use of linear biomechanical models (LIBR, left) provided a physics-based initialization that is accurate in many regions but degrades in others, especially in regions with large deformations (middle).  \ours learnd a deformation residual that selectively correct these poorly modeled areas, while remaining consistent with LIBR where its predictions are already reliable (right). 
Figure \ref{fig:error_wrt} (Left) shows that,  
while the linear biomechanical solutions (LIBR, blue) showed rapidly increasing errors at locations further away from sparse measurements, 
\ours (orange) was able to significantly reduce this rate of deterioration and correct linear deformation residuals at locations far away from sparse measurements, 
providing evidence for the long-range information transfer enabled by \ours's meta-learning formulation. 
Figure \ref{fig:error_wrt} (Left) shows that, 
while linear biomechnical deformation (blue) as expected was not able to model large deformations, 
\ours (oranging) was able to focus on correcting these large deformations, 
achieving its intended goal as resolving the residual in linear biomechanical deformations.

\section{Conclusion}

In this paper, we present \ours, 
a novel approach to image-to-physical registration 
via learning-to-learn 
a graph neural diffusion model of the residual of linear biomechanical deformation, 
using sparse intraoperative measurements as context samples. 
Our experimental results demonstrated the effectivess of 
this approach 
to resolve the errors in deformations predicted by linear biomechanics, 
especially in regions with large deformations and remote from intraoperative measurements. 
As a novel proof of concept, 
the current study is limited by evaluation on phatom datasets, 
the need to consider more diverse deformation patterns,  
and the need of an in-depth ablation of the various geometrical features that can further enhance the performance of \ours.
\bibliographystyle{splncs04}
\bibliography{bibliography}

@inproceedings{Finn2017MAML,
  title     = {Model-Agnostic Meta-Learning for Fast Adaptation of Deep Networks},
  author    = {Finn, Chelsea and Abbeel, Pieter and Levine, Sergey},
  booktitle = {Proceedings of the 34th International Conference on Machine Learning (ICML)},
  year      = {2017},
  volume    = {70},
  series    = {Proceedings of Machine Learning Research},
  pages     = {1126--1135}
}

@book{BonetWoodNonlinearContinuumMechanics,
  title     = {Nonlinear Continuum Mechanics for Finite Element Analysis},
  author    = {Bonet, Javier and Wood, Richard D.},
  publisher = {Cambridge University Press},
  year      = {2008}
}

@book{Belytschko2013NonlinearFEM,
  title     = {Nonlinear Finite Elements for Continua and Structures},
  author    = {Belytschko, Ted and Liu, Wing Kam and Moran, Brian},
  edition   = {2},
  year      = {2013},
  publisher = {Wiley}
}

@incollection{Meyer2003DDG,
  title     = {Discrete Differential-Geometry Operators for Triangulated 2-Manifolds},
  author    = {Meyer, Mark and Desbrun, Mathieu and Schr{\"o}der, Peter and Barr, Alan H.},
  booktitle = {Visualization and Mathematics III},
  publisher = {Springer},
  year      = {2003}
}

@article{abuhilal_southampton_2018,
  title = {The {{Southampton Consensus Guidelines}} for {{Laparoscopic Liver Surgery}}: {{From Indication}} to {{Implementation}}},
  shorttitle = {The {{Southampton Consensus Guidelines}} for {{Laparoscopic Liver Surgery}}},
  author = {Abu Hilal, Mohammad and Aldrighetti, Luca and Dagher, Ibrahim and Edwin, Bjorn and Troisi, Roberto Ivan and Alikhanov, Ruslan and Aroori, Somaiah and Belli, Giulio and Besselink, Marc and Briceno, Javier and Gayet, Brice and D'Hondt, Mathieu and Lesurtel, Mickael and Menon, Krishna and Lodge, Peter and Rotellar, Fernando and Santoyo, Julio and Scatton, Olivier and Soubrane, Olivier and Sutcliffe, Robert and Van Dam, Ronald and White, Steve and Halls, Mark Christopher and Cipriani, Federica and Van Der Poel, Marcel and Ciria, Ruben and Barkhatov, Leonid and {Gomez-Luque}, Yrene and {Ocana-Garcia}, Sira and Cook, Andrew and Buell, Joseph and Clavien, Pierre-Alain and Dervenis, Christos and Fusai, Giuseppe and Geller, David and Lang, Hauke and Primrose, John and Taylor, Mark and Van Gulik, Thomas and Wakabayashi, Go and Asbun, Horacio and Cherqui, Daniel},
  year = 2018,
  month = jul,
  journal = {Annals of Surgery},
  volume = {268},
  number = {1},
  pages = {11--18},
  issn = {0003-4932, 1528-1140},
  doi = {10.1097/SLA.0000000000002524},
  urldate = {2026-02-08},
  langid = {english}
}

@inproceedings{brunet_physicsbased_2019,
  title = {Physics-{{Based Deep Neural Network}} for {{Augmented Reality During Liver Surgery}}},
  booktitle = {Medical {{Image Computing}} and {{Computer Assisted Intervention}} -- {{MICCAI}} 2019},
  author = {Brunet, Jean-Nicolas and Mendizabal, Andrea and Petit, Antoine and Golse, Nicolas and Vibert, Eric and Cotin, St{\'e}phane},
  editor = {Shen, Dinggang and Liu, Tianming and Peters, Terry M. and Staib, Lawrence H. and Essert, Caroline and Zhou, Sean and Yap, Pew-Thian and Khan, Ali},
  year = 2019,
  pages = {137--145},
  publisher = {Springer International Publishing},
  address = {Cham},
  doi = {10.1007/978-3-030-32254-0_16},
  isbn = {978-3-030-32254-0},
  langid = {english}
}

@inproceedings{chamberlain_grand_2021,
  title = {{{GRAND}}: {{Graph Neural Diffusion}}},
  shorttitle = {{{GRAND}}},
  booktitle = {Proceedings of the 38th {{International Conference}} on {{Machine Learning}}},
  author = {Chamberlain, Ben and Rowbottom, James and Gorinova, Maria I. and Bronstein, Michael and Webb, Stefan and Rossi, Emanuele},
  year = 2021,
  month = jul,
  pages = {1407--1418},
  publisher = {PMLR},
  issn = {2640-3498},
  urldate = {2026-02-01},
  langid = {english}
}

@inproceedings{heiselman_comparison_2023,
  title = {Comparison Study of Sparse Data-Driven Soft Tissue Registration: Preliminary Results from the Image-to-Physical Liver Registration Sparse Data Challenge},
  shorttitle = {Comparison Study of Sparse Data-Driven Soft Tissue Registration},
  booktitle = {Medical {{Imaging}} 2023: {{Image-Guided Procedures}}, {{Robotic Interventions}}, and {{Modeling}}},
  author = {Heiselman, Jon S. and Collins, Jarrod A. and Ringel, Morgan J. and Jarnagin, William R. and Miga, Michael I.},
  year = 2023,
  month = apr,
  volume = {12466},
  pages = {150--161},
  publisher = {SPIE},
  doi = {10.1117/12.2655468},
  urldate = {2026-02-09}
}

@article{heiselman_intraoperative_2020,
  title = {Intraoperative {{Correction}} of {{Liver Deformation Using Sparse Surface}} and {{Vascular Features}} via {{Linearized Iterative Boundary Reconstruction}}},
  author = {Heiselman, Jon S. and Jarnagin, William R. and Miga, Michael I.},
  year = 2020,
  month = jun,
  journal = {IEEE Transactions on Medical Imaging},
  volume = {39},
  number = {6},
  pages = {2223--2234},
  issn = {1558-254X},
  doi = {10.1109/TMI.2020.2967322},
  urldate = {2025-04-09}
}

@article{miga_computational_2016,
  title = {Computational {{Modeling}} for {{Enhancing Soft Tissue Image Guided Surgery}}: {{An Application}} in {{Neurosurgery}}},
  shorttitle = {Computational {{Modeling}} for {{Enhancing Soft Tissue Image Guided Surgery}}},
  author = {Miga, Michael I.},
  year = 2016,
  month = jan,
  journal = {Annals of Biomedical Engineering},
  volume = {44},
  number = {1},
  pages = {128--138},
  issn = {1573-9686},
  doi = {10.1007/s10439-015-1433-1},
  urldate = {2026-02-08},
  langid = {english}
}

@inproceedings{pfeiffer_nonrigid_2020,
  title = {Non-{{Rigid Volume}} to {{Surface Registration Using}} a {{Data-Driven Biomechanical Model}}},
  booktitle = {Medical {{Image Computing}} and {{Computer Assisted Intervention}} -- {{MICCAI}} 2020},
  author = {Pfeiffer, Micha and Riediger, Carina and Leger, Stefan and K{\"u}hn, Jens-Peter and Seppelt, Danilo and Hoffmann, Ralf-Thorsten and Weitz, J{\"u}rgen and Speidel, Stefanie},
  editor = {Martel, Anne L. and Abolmaesumi, Purang and Stoyanov, Danail and Mateus, Diana and Zuluaga, Maria A. and Zhou, S. Kevin and Racoceanu, Daniel and Joskowicz, Leo},
  year = 2020,
  pages = {724--734},
  publisher = {Springer International Publishing},
  address = {Cham},
  doi = {10.1007/978-3-030-59719-1_70},
  isbn = {978-3-030-59719-1},
  langid = {english}
}

@inproceedings{wang_libr_2024,
  title = {{{LIBR}}+: {{Improving Intraoperative Liver Registration}} by~{{Learning}} the~{{Residual}} of~{{Biomechanics-Based Deformable Registration}}},
  shorttitle = {{{LIBR}}+},
  booktitle = {Medical {{Image Computing}} and {{Computer Assisted Intervention}} -- {{MICCAI}} 2024},
  author = {Wang, Dingrong and Azadvar, Soheil and Heiselman, Jon and Jiang, Xiajun and Miga, Michael and Wang, Linwei},
  editor = {Linguraru, Marius George and Dou, Qi and Feragen, Aasa and Giannarou, Stamatia and Glocker, Ben and Lekadir, Karim and Schnabel, Julia A.},
  year = 2024,
  pages = {359--368},
  publisher = {Springer Nature Switzerland},
  address = {Cham},
  doi = {10.1007/978-3-031-72089-5_34},
  isbn = {978-3-031-72089-5},
  langid = {english}
}

@article{yang_boundary_2024,
  title = {Boundary {{Constraint-free Biomechanical Model-Based Surface Matching}} for {{Intraoperative Liver Deformation Correction}}},
  author = {Yang, Zixin and Simon, Richard and Merrell, Kelly and Linte, Cristian A.},
  year = 2024,
  journal = {IEEE Transactions on Medical Imaging},
  pages = {1--1},
  issn = {1558-254X},
  doi = {10.1109/TMI.2024.3515632},
  urldate = {2025-03-17}
}
\end{document}